\def\year{2022}\relax
\documentclass[letterpaper]{article} % DO NOT CHANGE THIS
\usepackage{aaai22}  % DO NOT CHANGE THISx
\usepackage{times}  % DO NOT CHANGE THIS
\usepackage{helvet}  % DO NOT CHANGE THIS
\usepackage{courier}  % DO NOT CHANGE THIS
\usepackage[hyphens]{url}  % DO NOT CHANGE THIS
\usepackage{graphicx} % DO NOT CHANGE THIS
\usepackage{natbib}  % DO NOT CHANGE THIS AND DO NOT ADD ANY OPTIONS TO IT
\usepackage{caption} % DO NOT CHANGE THIS AND DO NOT ADD ANY OPTIONS TO IT
\DeclareCaptionStyle{ruled}{labelfont=normalfont,labelsep=colon,strut=off} % DO NOT CHANGE THIS
\usepackage{algorithm}
\usepackage{algorithmic}

\usepackage{newfloat}
\usepackage{listings}
\floatstyle{ruled}
\newfloat{listing}{tb}{lst}{}
\floatname{listing}{Listing}
\newcommand{\titleText}{Graph-based Neural Modules to Inspect Attention-based Architectures: \\
A Position Paper}

\title{\titleText}
\author {
    Breno W. Carvalho,{\textsuperscript{\rm 1}}
    Artur S. d'Avila Garcez,{\textsuperscript{\rm 2}}
    Lu{\'{\i}}s C. Lamb{\textsuperscript{\rm 3}}
    }

\affiliations {
    \textsuperscript{\rm 1} IBM Research, Rio de Janeiro, Brazil\\
    \textsuperscript{\rm 2} Department of Computer Science, City, University of London\\
    \textsuperscript{\rm 3} UFRGS, Porto Alegre, Brazil; MIT Sloan School of Management, Cambridge, MA\\
    brenow@ibm.com,  a.garcez@city.ac.uk, lamb@inf.ufrgs.br
}

\begin{document}

\maketitle

\begin{abstract}
Encoder-decoder architectures are prominent building blocks of state-of-the-art solutions for tasks across multiple fields where deep learning (DL) or foundation models play a key role. Although there is a growing community working on the provision of interpretation for DL models as well as considerable work in the neuro-symbolic community seeking to integrate symbolic representations and DL, many open questions remain around the need for better tools for visualization of the inner workings of DL architectures. In particular, encoder-decoder models offer an exciting opportunity for visualization and editing by humans of the knowledge implicitly represented in model weights. %Currently, it is usually necessary to change the training dataset to correct the model's underlying understanding of the text. 
In this work, we explore ways to create an abstraction for segments of the network as a two-way graph-based representation. Changes to this graph structure should be reflected directly in the underlying tensor representations. Such two-way graph representation enables new neuro-symbolic systems by leveraging the pattern recognition capabilities of the encoder-decoder along with symbolic reasoning carried out on the graphs. The approach is expected to produce new ways of interacting with DL models but also to improve performance as a result of the combination of learning and reasoning capabilities.
%
%Encoder-decoder architectures are prominent building blocks or construction patterns of some state-of-the-art solutions for tasks across multiple fields dominated by deep learning or foundation models.
\end{abstract}

%%% INTRODUCTION
\section{Introduction}
\begin{figure}[ht!]
    \centering
    \includegraphics[width=.8\columnwidth]{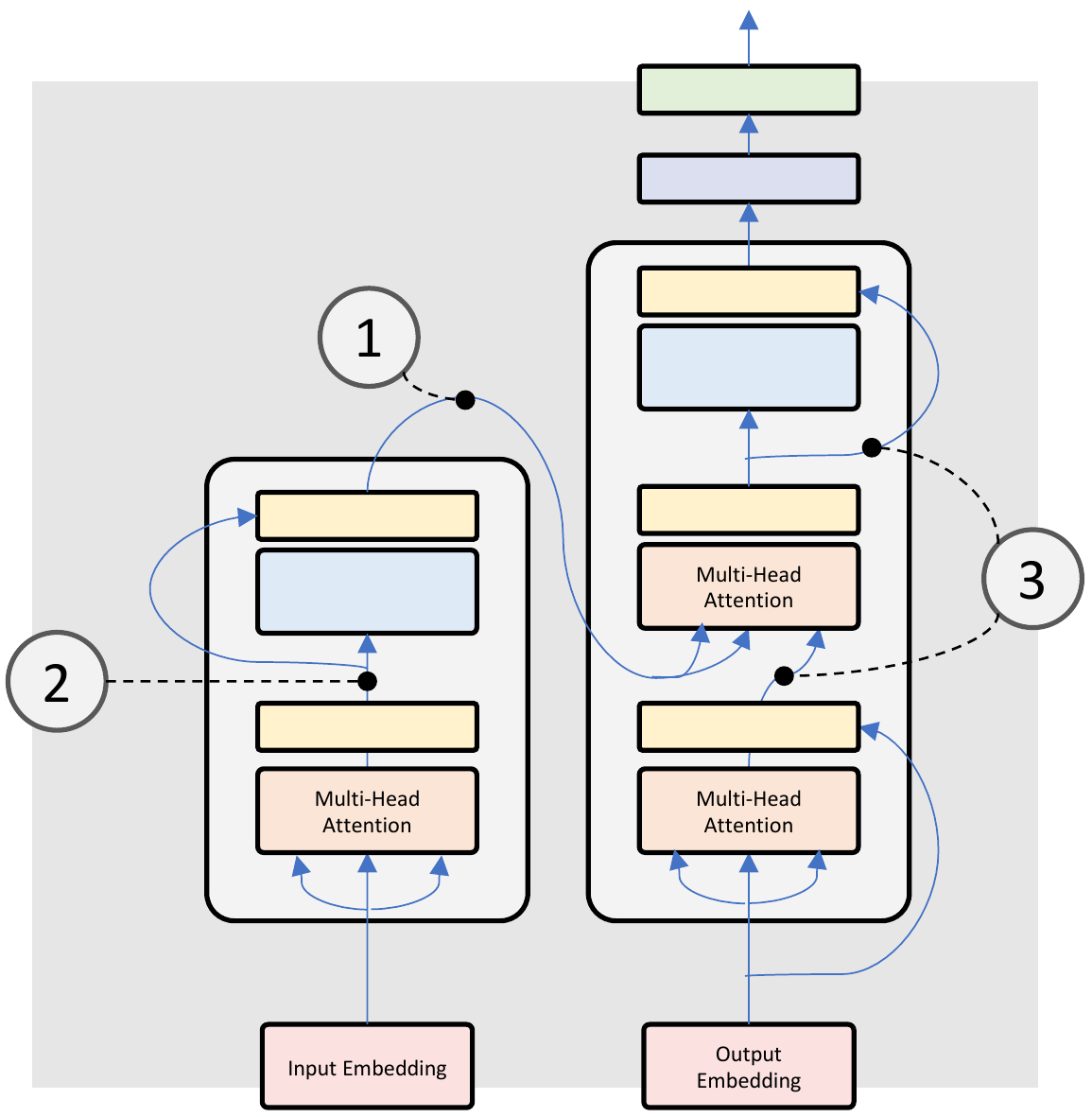}
    \caption{Standard transformer architecture example, adapted from \cite{NIPS2017_attention} with a few possible entry points for introspection modules. Each one of those prospective entry points carries information about the model state and context.
        In 1) the decoder part of the transformer uses the embedded context information from the encoder to generate the output sequence in the transduction task. In
        2) after the masked multi-head attention module of the \emph{encoder} component to inspect which concepts are more strongly related than others in the construction of contextual information.
        In 3) information from the masked multi-head attention modules of the \emph{decoder} component is used to inspect which concepts are more relevant to processing the context in light of the current output state.
    }
    \label{fig:transformer-entry-points}
\end{figure}

We live in hustling times for Artificial Intelligence (AI) research with many optimistic perspectives from researchers from Computer Science and other areas. We are entering an era where our models not only can generalize from examples of a given task, but given the appropriate context and conditions, they can generalize across different tasks, indicating an emergent phenomenon that is almost impossible to predict. Such models, notably deep learning (DL) ones are powerful and influential, yet the current drawbacks in design and reliability are now obvious. Those models, like fun-house mirrors, can distort aspects of their training data and generate false affirmations even though they were fed with the correct facts necessary for the answer. They can also lean toward unethical biases and make unsupported claims. It is therefore important to be able to edit a model's answer and context to first understand the supporting facts or assumptions behind a given answer and second, correct or provide new contextual  information so as to increase trust in the model's outputs.

One approach in this direction is to create introspection entry points into the models so as to \textit{peak} into the operations of an intermediate layer (as depicted in Figure ~\ref{fig:transformer-entry-points}), interacting directly with it via graph editing. Those introspection entry points are neural layers not aimed to contribute to performance in any single task but to add an extra layer to the developer's interaction with the model by providing an abstraction to inspect specific parts of the network.
There are already a few relevant methods in the literature, some of which are listed in the survey \cite{GNNmeetNeSy} and use graph representations as part of the network structure itself. At the same time, the field of neuro-symbolic (NeSy) AI research investigates ways of merging explicit symbolic knowledge with many such sub-symbolic neural approaches \cite{NesySurvey}.

At the Robert S. Engelmore Memorial Lecture, during the AAAI Conference on Artificial Intelligence, New York, February 10\textsuperscript{th}, 2020, in a talk entitled \emph{The Third AI Summer}, Henry Kautz presented a taxonomy for NeSy models, with six main categories \cite{Kautz22}.
Adding those inspection nodes, which ideally should not harm model performance at the target task, falls into the \emph{Neuro;~Symbolic} category in Kautz's taxonomy. This category includes systems where a neural component and a symbolic component mutually share information, each with a different goal. In this way, the inspection module can help edit the implicit knowledge stored in the DL model and serve even as a debugging tool. One possible approach to building such inspection modules is based on Graph Convolutional Neural Networks (GNN) \cite{liu2020gnn_book, wu2020gnnsurvey}. By combining inspection with the model one can perform a secondary task of internal state-representation of the model~\cite{GNNmeetNeSy}.
As DL models grow in size and complexity, we believe that those introspection tools in the network may allow for a more in-depth understanding of such models. This raises the question of the form of introspection and extensions thereof that might be expected to produce a better understanding of DL models.

%===================================================================================================================================================
\section{Short-term Objectives}

Our immediate objective is to design and develop an initial tool set for interpreting encoded knowledge in deep learning (large) language models and editing it. We aim to add components to infer a graph structure from specific tensors, so-called entry points, in the model's architecture. In this way, a natural step is to evaluate the feasibility of implementing different approaches for those graph entry points (see Figure~\ref{fig:encoder-graph-decoder}). We call entry points compact and non-sparse tensors along the network pipeline that can also be connected to a GNN to produce a graph abstraction of this stage of the overall network. For now, we consider only DL architectures based on the overall encoder-decoder pattern. Those are usually models used in transduction tasks, which make up a vast class of DL models. Our short-term goals are:

\paragraph[Graph constraints]{To compare different alternatives for implementing graph constraints without significantly reducing performance.}
Creating graph constraints to specific entry points means ensuring that the underlying tensor always has a meaningful compact graph representation if the overall model was fed with valid input vectors.
To this end, we contemplate adopting a few alternatives, such as having an auxiliary loss function that will signal whether the encoder was able to create a valid graph. It will also signal if the decoder can generate the expected output. %A few examples of this goal are present in the literature, such as [REFS here].
Another approach is to train a GNN decoder and apply it to the output of the encoder part. Aiming at language tasks supports both kinds of strategies since there are extensive Language Resources that can be used as an initial graph structure.

\paragraph[Vocabulary and semantics]{To  find semantic representations for the edges and nodes.}
A representation of one embedding without any semantics associated with it might be as daunting as looking at the model itself. We
seek to adapt existing work in the literature to be able to create a representation with more meaningful names for entity nodes and connections. %Some approaches that we might take inspiration from are proposed in [REF (with static ontology)] and [REF (from the text most important tokens or entities)]

\paragraph[Human editing and interaction with the model]{To be able to edit the model to increase performance, remove biases or add new knowledge.} 
Finally, any editing of the graph representation should be reflected in the encoder's weights and influence the final output of the model, ideally in a way that suggests a causal relationship.
This enables us to interact actively and edit context within the model to ensure that the main component of a given output is considered to be relevant and correct by an expert in the domain of interest.

\paragraph[Local scope]{To Design a Method for generating interpretations in a local scope.}
In the short term, we are  concerned with \emph{local interpretations} where we interpret the model components' states and context while processing a specific input. This approach is already insightful for a series of use cases and might be the stepping-stone for developing a global interpretation.

Achieving these goals will require the development of a simple but versatile tool-set that might be useful to the community for inspecting and editing models in diverse research or industrial areas (similarly to Grad-Cam~\cite{selvaraju2017gradcam} and Bertology~\cite{rogers2020Bertology} and many other helpful visualization tools found to be relevant by DL practitioners).

%===================================================================================================================================================
\section{Long-term Objectives}
Our main goal is to be able to inspect large deep learning language models through different entry points in such a way as to increase trust in their output. Transformer or transformer-inspired architectures dominate the state-of-the-art of such language models.
Considering that most transformer architectures follow an encoder-decoder model, fulfilling the above short-term objectives should provide an entry point into inspecting the inner state of the model and its context.
We conjecture that having a few editable graph layers attached to a transformer of considerable size will enable an expert to study the relationships of the many abstraction levels one might find by inspecting those layers. Our long-term goals are: 

\begin{figure}[htb] 
    \centering
    \includegraphics[width=.8\columnwidth]{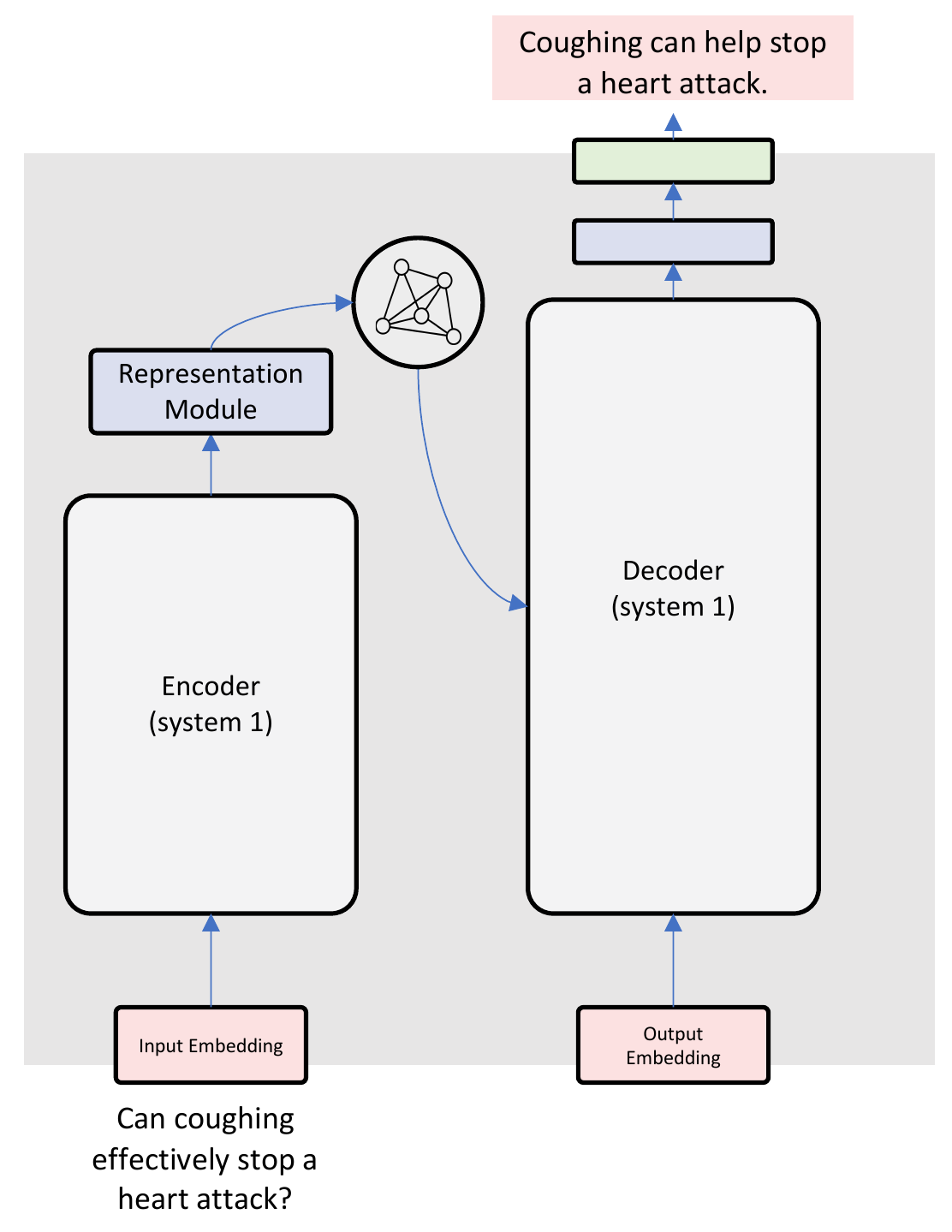}
    \caption{Encoder-decoder example with an entry point for an interpretation module. One might explore multiple representation languages depending on the resources available. In this case, the model outputs an incorrect answer for a question from the TruthfulQA dataset, \cite{TruthfulQA}, due to incomplete context information, that we hope to make explicit with the graph representation.}
    \label{fig:encoder-graph-decoder}
\end{figure}

\begin{figure}[htb] 
    \centering
    \includegraphics[width=.8\columnwidth]{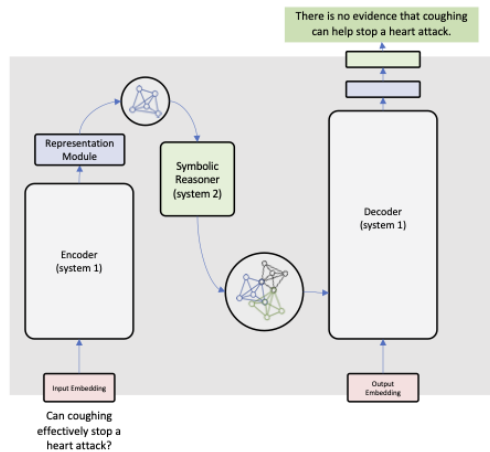}
    % \vspace{.3in}
    % \centerline{\fbox{Overal diagram \newline \fbox{GNN Fast and Slow} \fbox{?}}}
    % \vspace{.3in}
    \caption{Illustrative representation of neuro-symbolic reasoning to complement the pattern learning capabilities of attention-based models. A GNN entry point is interpreted as an AI system 2 component via the application of a symbolic reasoner.}
    \label{fig:think_fast-slow}
\end{figure}

\paragraph[Interacting and editing the models]{To have a method to enable experts to visualize, edit and interact directly with the language models.} One might consider exploring multiple entry points.
Manually editing context or knowledge in the network through a graph interface linked to different layers might provide much richer insight. 
Possibilities for graph-entry points are depicted in Figure~\ref{fig:transformer-entry-points}. The goal is to select tensors that are most descriptive of the model's current state. One can also use this method with visualization methods such as \cite{rogers2020Bertology}.

\paragraph[Systems 1 and 2]{To design a new family of neurosymbolic models based on systematic interactions with fragments of the underlying architecture.}

One might also leverage the power of symbolic reasoners to perform expansion and inference on the graph context representation to fix or enhance the model's output. This approach is much aligned with a fast-slow AI perspective, where the learning of the language model represents the fast, pattern-matching, system 1, and the slower, graph-based reasoning represents system~2 (see Figure~\ref{fig:think_fast-slow}). In this figure, we illustrate that a symbolic expansion of the graph representation of specific parts of the network might be used to provide the model with enough information to go from a wrong answer, as depicted in Figure~\ref{fig:encoder-graph-decoder}, to the correct answer (see  Figure~\ref{fig:think_fast-slow}).

\paragraph[Alternative representations]{To explore alternative representations of the entry points that are most useful in a specific context.}
As we explore larger language models we might find that different entry points might benefit from different representation formats. It might even be the case that reusing smaller transformers will allow us to have natural language representation for some model entry points.
An analogy would be when humans perform complex tasks and mentally talk to themselves while performing the task. Another class of representation alternatives to be explored consists of hyper-graphs that might have a more suitable role in explaining the network's inference mechanism since they are able to convey recursive levels of abstraction in the information being represented.

\paragraph[Data-driven semantics]{To develop a method where the training data gives the representation semantics and not necessarily by external language resources.}
In an ideal scenario, the only information available during training should be sufficient to build meaningful representations. Even with available linguistic resources, expanding the entry point representation with tokens and concepts from the training data could allow the use of a more robust and extensible representational language such as a graph, hypergraph, or natural language.

\paragraph[Pre-trained zoo]{To develop a pre-trained zoo of entry point representations for specific architectures that can be further specialized.}
It should be possible to use a library of visualization modules pre-trained on commonplace text corpora as a baseline, and specialize it with specific domain corpora. This process might accelerate the inspection process and also diminish the learning curve of experts from different domains on how to use those tools.
At this point, we are aiming at modularity. We intend not only to have a methodology but also specific modules that experts from diverse backgrounds can reuse.

\paragraph[Local and global interpretation scope]{To define a method comprising both local and global interpretation scope.}
In opposition to local scope interpretations, \emph{global interpretations} aim at providing insight into the overall behavior of the model considering an entire task. For instance, understanding what parts of a large language model are responsible for summing two numbers or performing sentiment analysis.

To achieve those objectives, one should build a robust methodology to edit and interact with large language models at multiple levels of abstraction processed by the model. This methodology might enable further insight into the inner workings of such models.

%===================================================================================================================================================
\section{Introspection Modules}
%Intro
    Adding introspection modules to current deep learning architectures has the potential to enable better interpretability, but also unlock new ways to edit and condition those models.
    As discussed in detail by this joint initiative \cite{foundation_models_risks_opportunities2021}, large-scale language models (also called foundation models for their re-usability across multiple tasks) pose serious research and societal challenges. We discuss a few of them in this paper's \emph{Challenges} section.

\subsection{Fairness and accountability goals}
%Fairness
     Although somewhat hard to quantify, fairness and accountability are important to acknowledge in what concerns trust in deep learning models of far-reaching decisions. Unfairness is characterized in \cite{du2020fairness} into two classes: prediction outcome discrimination, and prediction quality disparity. An example of the first class is when a model predicts an unfavorable treatment of a demographic group, such as bias against women. The second class refers to a model performing poorly for a determined group of individuals. Both cases raise the need to flag such models and correct them. Accountability might be understood as a capability for \emph{post-hoc} inspection of the model in order to make it available for auditing the causes of the model's outcome \cite{chakraborty2017interpretability}.
     
%Interpretability.
    Following \cite{foundation_models_risks_opportunities2021}, Section 4.11, which discusses the concept of interpretability of foundation models, we contemplate three categories of interpretability: \emph{what} (the model's limits and capabilities), \emph{why} (finding what in the data set is responsible for the model's output), and \emph{how} (gaining an understanding of how certain parts and mechanisms of the model impact the output). In this work, we focus on the \emph{why} and \emph{how} interpretability questions.

As well as being able to interpret those models, one should also seek to obtain better tools to assist us in accelerating the construction or adaptation of new models.

\subsection{Model editing and design goals}
% %Human editing and interaction
%     Iteratively human-in-the-loop knowledge editing.
    
%Composability
    By understanding individual model mechanisms, we seek to build a compositional understanding of the complex behavior of a foundation model.
    A possible way is to interconnect or build larger models from pre-trained components and
    reduce societal risks by building more reliable and interpretable models through ``debugging" tools. 

%AG: Too strong unsupported claim! Since we seek to develop a methodology that ultimately induces a better understanding of those large models, it may even help us find new research directions to assist the scientific community in the long-term pursuit of general AI. 

%===================================================================================================================================================
\section{Challenges}\label{sec:challenges}
Creating those representation/introspection modules raises several challenges. Some of these, such as the visual interpretation of self-attention heads, have been addressed to some extent in the literature. Others remain as yet to be explored.

The \emph{why} and \emph{how} interpretability questions from \cite{foundation_models_risks_opportunities2021}, as mentioned in the previous section, are essential for the understanding of large language models, but face many open challenges:

\subsection[Describing model's behavior]{Challenges on describing \emph{why} (model behavior) and \emph{how} (model mechanisms)}
Studying the influence of the model's input might be performed by carefully studying how the behavior of the model changes with changes in the input. The ability to peek inside the model also might be used to provide insight into which parts of the input are more relevant to the model. The task of generating interpretations for the inner workings of a language model poses further challenges:

    \paragraph[Multiplicity]{1) It is not clear how the components of those language models are interconnected.}
     Although the overall design of a model's architecture may be clear, the actual weights of such units have emergent behavior that might be fairly hard to predict. A central question to a fair representation of the mechanisms inside a language model is that of to what extent those mechanisms behave as one coherent model or as many models. As the model learns to generalize across multiple tasks, it might be the case that its components find weight sub-spaces that virtually correspond to many sub-models, possibly one for each task. It isn't clear yet to which point in this spectrum current foundation models belong. This lack of understanding and clarity leads to our next challenge.

    \paragraph[Local vs Global interpretations]{2) Local and global interpretation methods might have pitfalls related to the complexity of such language models.}
     Given the highly complex behavior of those models and their emergent properties, it is unwise to jump to conclusions based on a single global or local interpretation method, even though there are both local\cite{fong2017meaningfulPertubation, koh2017InfluenceFunctions} and global\cite{lapuschkin2019unmaskingCleverHans, thomas2019neurimageDL} methods for providing insights into task-specific models. For instance, considering that those large models might find weight sub-spaces for distinct tasks it might be the case that studying the model for a few of those sub-tasks doesn't provide a fair picture of its behavior in other tasks or even for other similar inputs.

     \paragraph[Unclear coupling of semantically related components]{3) The inter-relationship of different components in the model might be counter-intuitive or hard to determine.}
     It is not clear whether semantically related mechanisms, e.g. the parts of a language model responsible for summing up two numbers, are the same as the ones used to perform a related task, such as e.g. summing two numbers expressed as numeral nouns.

%\subsection[Describing model's mechanisms]{Challenges on \emph{how} --- Describing model's mechanisms}

\subsection[Introspection modules]{Challenges about the introspection modules themselves}
Besides the inherent challenges of deriving interpretations from huge language models, there are also intrinsic points to be evaluated in the design of the introspection model. There are questions and decisions to be made about the desired representational language and format, and there are also considerations about the architecture of those modules and their refinement procedure.

    \paragraph[Vocabulary and semantics]{1) Creating meaningful interpretations assumes an underlying vocabulary and semantics.}
     This can be thought of as the symbol grounding problem, i.e. the definition of a vocabulary for the representation generated by the introspection modules. This might be achieved through the use of an underlying ontology that will constrain the interpretation space into a set of predefined concepts and vocabulary. It remains unclear how the semantics of the ontology will be transferred to the interpretation.
     It might be achieved as a secondary task happening during the training of the language model that we want to inspect.
    
    \paragraph[Too complex interpretations]{2) Different abstraction levels must be considered in order to avoid overly complicated and misguiding interpretations.}
     This happens because the size of a given interpretation may be prohibitive, both in terms of meaningful visualization and interpretability. To circumvent this one needs to employ different levels of abstraction, although those model layers closer to the output layers seem to pertain to more abstract features of the data set. Walking back and forth between those different levels of abstraction reinforces the need to guarantee consistency in the interpretations' vocabulary from different entry points.
    
    \paragraph[Consistency and reliability]{3) Interpretations should be consistent and reliable across a specific domain or data set.}
     This is not an easy requirement to enforce. It might be the case that given the complexity of those models, consistency can be enforced only partially for most of the time. We also need to consider consistency across related input or context. For instance, it should not be the case that two very similar text prompts result in very different interpretations when looking at the same entry point in the model. A possible approach is to anchor those graph representations on concepts and constructs from public linguistic resources, such as FrameNet~\cite{baker1998framenet}, VerbNet~\cite{palmer2004verbnet}, WordNet~\cite{1995wordnet} and Propbank~\cite{palmer2005propbank}, all connected by the SemLink project~\cite{palmer2009semlink}.
     The process of generating a consistent interpretation must also be subject to auditing to ensure the faithfulness of the entire system.
    
    \paragraph[Introspection modules might be black-box components]{4) Considering that the introspection module can be viewed as a black-box component, one must devise metrics of trustworthiness specific to it.}
    The GNN or other DL model used to generate representations from parts of the language model is also a black-box module that might pose itself some fairness questions \cite{jacovi2020faithfulNLP}. 
    We hope that it will prove feasible to keep those models relatively small. It is important to make sure that relying on those modules will be much less complex than the model being studied. In such case, existing visualization methods such as \cite{rogers2020Bertology, shi2021visualizingGCN} may suffice to audit them and improve trust.

In summary, the goal of adding a human-editable and interpretable representation of context and the state of specific parts of a language model has many open questions that to be solved will require drawing contributions from different areas of expertise.
One way to face some of the challenges outlined here is to combine approaches that use both linguistic resources to provide the needed representation building blocks (e.g. concepts and relations) and to use tokens and concepts from the training data set as building blocks.

%===================================================================================================================================================

\section{Outlook}
%What is ready in the literature: datasets, approaches, resources, communities etc
As mentioned in previous sections, this kind of neural-symbolic work draws from a diverse mixture of research areas in AI and Computer Science more generally.

\subsection[Peeking inside DL models]{Peeking inside DL models:}
While characterizing and describing the behavior of a deep learning model, one might consider the data set and the predicted outputs, or also the state variations of the models. It is also possible to build interpretations about the model behavior on specific data samples or trying to understand how it behaves in a broader sense when performing a particular task.

Local scope interpretation methods target insight into the model's behavior on specific inputs or contrasting a given set of data points. Examples include \cite{koh2017InfluenceFunctions}, an  approach that uses influence functions from robust statistics to understand how perturbations in the training set influence the model behavior; \cite{fong2017meaningfulPertubation} uses the concept of interpretations as meta-predictors, i.e. one can use an explanation to predict the output of a model, and they instantiate this approach using image classification masks.
By contrast, global methods are used to describe general aspects and behaviors of the model. Interesting examples include \cite{lapuschkin2019unmaskingCleverHans} and \cite{thomas2019neurimageDL}.
Both local and global methods treat the language models as black-box objects to be understood through their inputs and outputs. 

Other than looking at the training set and the models' outputs, another class of interpretation methods that is more aligned with the work proposed here, consists of models that extract information from the model weights. This includes GradCam \cite{selvaraju2017gradcam}, which does backward propagation of gradients to estimate which pixels of the input where more relevant for a Convolutional Network output; \cite{rogers2020Bertology} provides a tool-set for visualizing the weights of the self-attention heads of BERT-like models.
Although those methods can be very insightful when analyzing task-specific models, considering that current language models might have distinct behavior across different tasks, it is unclear how insightful those methods could be in this scenario. Most likely, the designers and analysts of such models would need to employ multiple methods of interpretation and visualization to have a fair understanding of the models' behavior.

As an alternative to the line of work described in this paper, where we propose the use of external (and hopefully auditable) modules to peak into the language models, one can train the model itself to generate an explanation for its output \cite{elton2020selfexplanation}.
This approach faces some skepticism since language models can generate plausible, although untrue, outputs and nothing prevents this behavior from extending to the model's self-explanations.

\subsection{Graph deconvolution and Neurosymbolic AI}
We regard GNNs and neuro-graph algorithms in general as a promising general approach for both local and global interpretation.

Graph convolution is defined by analogy to convolutional layers over Euclidean data \cite{scarselli2008GNN, sanchez2021GentleGNN} and it is an important tool to infuse deep learning models with relational knowledge \cite{GNNmeetNeSy}. Conversely, we conjecture that using specific decoder units to extract graph representations from the entry point tensors (even if having to condition the training of the underlying language model) might be a worthwhile approach to obtain meaningful interpretations from language models.

We might consider each introspection module as a simple decoder unit that maps the encoded embedding into the interpretation space. As mentioned in the Challenges section, one might initially use linguist resources conditioned on the sentence processed by the language model to assemble the vocabulary used to form the interpretations. A possible approach is to use heuristics to get concepts from the language model prompt and build a simple graph, even combining it with a syntax or dependency tree to have an initial graph to use and evaluate the decoder unit used as an introspection module. In this sense, there are a few approaches in the literature that one might draw inspiration from. For instance, there is work on graph encoders  \cite{hamilton2017GraphRepresentation}, and also deconvolutional graph networks \cite{li2021GraphDeconvolution} to derive a graph from the entry point embedding. At this stage, it might be unclear how to define the appropriate abstraction level of the interpretation.

Instead of building the modules as decoder units trained on a heuristic graph, another approach is to adapt variational graph autoencoders \cite{kipf2016variationalGraph}. A variation of the encoding part could output a Graph Convolutional Network-like embedding, and the decoder would try to retrieve the original input of the language model (assuming that the entry point embedding might encode a smooth topological representation of concepts). This assumption is not guaranteed to hold, but it is an interesting issue to investigate. One might use deconvolution networks \cite{li2021GraphDeconvolution} to infer a graph without prior conditioning. The intuition behind this rationale is that this deconvolution approach would create representations without any supervision, which might be too complicated or obscure to be used as interpretation.

As stated in our short and long-term objectives, inferring an arbitrary graph from those entry points is not enough for our interpretation and editing purposes. Those graphs also need to have a clear semantics that correlates with the model's output and with commonsense and domain knowledge as well.

One might wonder why bother with graph representations and not try to go directly to building natural language interpretations. We conjecture that natural language will lack explicit restrictions that are needed for clarity of a given interpretation and also that the resulting natural language introspection module would need to be almost as complex as the language model itself, thus defeating its original purpose.

There are several neurosymbolic methods that aim to infuse the deep learning model architectures with symbolic or logical structures \cite{NesySurvey}, including logic tensor networks (LTN) \cite{serafini2016LTN} and logical neural networks (LNN) \cite{riegel2020LNN}.
They can also be used as parts of the introspection modules described here in order to make DL models more auditable and reliable.

% %===================================================================================================================================================
% \section{Approach?}

% Knowledge graph literature.
% Automatic KBC.
% Convolutional Graph Networks and GraphSage. Works to encode a graph works to create a graph from a vector representation.

%===================================================================================================================================================
\section{Conclusion}

Large language models are becoming state-of-the-art in many tasks and fields, and as they grow in popularity, they also grow in complexity. Those models usually follow a transduction encoder-decoder pattern based on self-attention mechanisms that are repeated multiple times to a point where they are so complex that surprising behavior emerges from them. In this paper, we state the need for inspection methods to both characterize and describe those models' behaviors. Despite the consolidated literature on deep learning interpretability, the community still needs to build different tools to inspect different aspects of those models. We believe that graph-based algorithms, in particular, deconvolution graph networks and variational graph auto-encoders might be key to generating formal yet flexible interpretations of parts of such large language models.
%Wide applicability and open problems.

%\paragraph[Blooming field and opportunities.]{}
Interpretability of language models is a blooming field as those models keep growing and reaching multiple uses in society. There is a vast literature to support this field, but there is a flagrant need for new methods and approaches to deal with such models' high generalizability.
Being able to audit and edit aspects of the model during development and deployment should allow the community to correct flawed inferences performed by those models and adjust to counter certain unethical biases. Thus inspection tools have the potential for tremendous impact if they can help us develop models we can trust and rely on to shape the further development of AI.

%\paragraph[Objectives]{}
We envision a roadmap for building tools with which multidisciplinary communities can start inspecting small transduction models based on self-attention, and progressively scale up to large-scale language models. By doing so, those communities are empowered to shape the widely spread use cases of such models in society. To this end, we hope to start a conversation to unify our efforts, lower the threshold for other machine learning researchers to join us, and bring these communities closer together with a common language.

% Use \bibliography{yourbibfile} instead or the References section will not appear in your paper

 \section{Acknowledgments}
  We thank Viviane Torres, Sandro Rama Fiorini, Emílio Ashton Brasil, and Renato Cerqueira for insightful conversations on the topic. Luis Lamb was supported in part by CAPES and CNPq, Brazil.
 \bibliography{references}

\begin{thebibliography}{31}
\providecommand{\natexlab}[1]{#1}

\bibitem[{Baker, Fillmore, and Lowe(1998)}]{baker1998framenet}
Baker, C.~F.; Fillmore, C.~J.; and Lowe, J.~B. 1998.
\newblock The berkeley framenet project.
\newblock In \emph{COLING 1998 Volume 1: The 17th International Conference on
  Computational Linguistics}.

\bibitem[{Bommasani et~al.(2021)Bommasani, Hudson, Adeli, Altman, Arora, von
  Arx, Bernstein, Bohg, Bosselut, Brunskill, Brynjolfsson, Buch, Card,
  Castellon, Chatterji, Chen, Creel, Davis, Demszky, Donahue, Doumbouya,
  Durmus, Ermon, Etchemendy, Ethayarajh, Fei{-}Fei, Finn, Gale, Gillespie,
  Goel, Goodman, Grossman, Guha, Hashimoto, Henderson, Hewitt, Ho, Hong, Hsu,
  Huang, Icard, Jain, Jurafsky, Kalluri, Karamcheti, Keeling, Khani, Khattab,
  Koh, Krass, Krishna, Kuditipudi, and
  et~al.}]{foundation_models_risks_opportunities2021}
Bommasani, R.; Hudson, D.~A.; Adeli, E.; Altman, R.; Arora, S.; von Arx, S.;
  Bernstein, M.~S.; Bohg, J.; Bosselut, A.; Brunskill, E.; Brynjolfsson, E.;
  Buch, S.; Card, D.; Castellon, R.; Chatterji, N.~S.; Chen, A.~S.; Creel, K.;
  Davis, J.~Q.; Demszky, D.; Donahue, C.; Doumbouya, M.; Durmus, E.; Ermon, S.;
  Etchemendy, J.; Ethayarajh, K.; Fei{-}Fei, L.; Finn, C.; Gale, T.; Gillespie,
  L.; Goel, K.; Goodman, N.~D.; Grossman, S.; Guha, N.; Hashimoto, T.;
  Henderson, P.; Hewitt, J.; Ho, D.~E.; Hong, J.; Hsu, K.; Huang, J.; Icard,
  T.; Jain, S.; Jurafsky, D.; Kalluri, P.; Karamcheti, S.; Keeling, G.; Khani,
  F.; Khattab, O.; Koh, P.~W.; Krass, M.~S.; Krishna, R.; Kuditipudi, R.; and
  et~al. 2021.
\newblock On the Opportunities and Risks of Foundation Models.
\newblock \emph{CoRR}, abs/2108.07258.

\bibitem[{Chakraborty et~al.(2017)Chakraborty, Tomsett, Raghavendra, Harborne,
  Alzantot, Cerutti, Srivastava, Preece, Julier, Rao
  et~al.}]{chakraborty2017interpretability}
Chakraborty, S.; Tomsett, R.; Raghavendra, R.; Harborne, D.; Alzantot, M.;
  Cerutti, F.; Srivastava, M.; Preece, A.; Julier, S.; Rao, R.~M.; et~al. 2017.
\newblock Interpretability of deep learning models: A survey of results.
\newblock In \emph{2017 IEEE smartworld, ubiquitous intelligence \& computing,
  advanced \& trusted computed, scalable computing \& communications, cloud \&
  big data computing, Internet of people and smart city innovation
  (smartworld/SCALCOM/UIC/ATC/CBDcom/IOP/SCI)}, 1--6. IEEE.

\bibitem[{d'Avila Garcez and Lamb(2020)}]{NesySurvey}
d'Avila Garcez, A.; and Lamb, L.~C. 2020.
\newblock Neurosymbolic {AI:} The 3rd Wave.
\newblock \emph{CoRR}, abs/2012.05876.

\bibitem[{Du et~al.(2020)Du, Yang, Zou, and Hu}]{du2020fairness}
Du, M.; Yang, F.; Zou, N.; and Hu, X. 2020.
\newblock Fairness in deep learning: A computational perspective.
\newblock \emph{IEEE Intelligent Systems}, 36(4): 25--34.

\bibitem[{Elton(2020)}]{elton2020selfexplanation}
Elton, D.~C. 2020.
\newblock Self-explaining AI as an alternative to interpretable AI.
\newblock In \emph{International conference on artificial general
  intelligence}, 95--106. Springer.

\bibitem[{Fong and Vedaldi(2017)}]{fong2017meaningfulPertubation}
Fong, R.~C.; and Vedaldi, A. 2017.
\newblock Interpretable explanations of black boxes by meaningful perturbation.
\newblock In \emph{Proceedings of the IEEE international conference on computer
  vision}, 3429--3437.

\bibitem[{Hamilton, Ying, and Leskovec(2017)}]{hamilton2017GraphRepresentation}
Hamilton, W.~L.; Ying, R.; and Leskovec, J. 2017.
\newblock Representation learning on graphs: Methods and applications.
\newblock \emph{arXiv preprint arXiv:1709.05584}.

\bibitem[{Jacovi and Goldberg(2020)}]{jacovi2020faithfulNLP}
Jacovi, A.; and Goldberg, Y. 2020.
\newblock Towards faithfully interpretable NLP systems: How should we define
  and evaluate faithfulness?
\newblock \emph{arXiv preprint arXiv:2004.03685}.

\bibitem[{Kautz(2022)}]{Kautz22}
Kautz, H.~A. 2022.
\newblock The Third {AI} Summer: {AAAI} Robert S. Engelmore Memorial Lecture.
\newblock \emph{{AI} Mag.}, 43(1): 93--104.

\bibitem[{Kipf and Welling(2016)}]{kipf2016variationalGraph}
Kipf, T.~N.; and Welling, M. 2016.
\newblock Variational graph auto-encoders.
\newblock \emph{arXiv preprint arXiv:1611.07308}.

\bibitem[{Koh and Liang(2017)}]{koh2017InfluenceFunctions}
Koh, P.~W.; and Liang, P. 2017.
\newblock Understanding black-box predictions via influence functions.
\newblock In \emph{International conference on machine learning}, 1885--1894.
  PMLR.

\bibitem[{Lamb et~al.(2020)Lamb, d'Avila Garcez, Gori, Prates, Avelar, and
  Vardi}]{GNNmeetNeSy}
Lamb, L.~C.; d'Avila Garcez, A.~S.; Gori, M.; Prates, M. O.~R.; Avelar, P.
  H.~C.; and Vardi, M.~Y. 2020.
\newblock Graph Neural Networks Meet Neural-Symbolic Computing: {A} Survey and
  Perspective.
\newblock In \emph{{IJCAI} 2020}, 4877--4884. ijcai.org.

\bibitem[{Lapuschkin et~al.(2019)Lapuschkin, W{\"a}ldchen, Binder, Montavon,
  Samek, and M{\"u}ller}]{lapuschkin2019unmaskingCleverHans}
Lapuschkin, S.; W{\"a}ldchen, S.; Binder, A.; Montavon, G.; Samek, W.; and
  M{\"u}ller, K.-R. 2019.
\newblock Unmasking Clever Hans predictors and assessing what machines really
  learn.
\newblock \emph{Nature communications}, 10(1): 1--8.

\bibitem[{Li et~al.(2021)Li, Li, Liu, Yu, Li, and
  Cheng}]{li2021GraphDeconvolution}
Li, J.; Li, J.; Liu, Y.; Yu, J.; Li, Y.; and Cheng, H. 2021.
\newblock Deconvolutional networks on graph data.
\newblock \emph{Advances in Neural Information Processing Systems}, 34:
  21019--21030.

\bibitem[{Lin, Hilton, and Evans(2021)}]{TruthfulQA}
Lin, S.; Hilton, J.; and Evans, O. 2021.
\newblock TruthfulQA: Measuring How Models Mimic Human Falsehoods.
\newblock \emph{CoRR}, abs/2109.07958.

\bibitem[{Liu and Zhou(2020)}]{liu2020gnn_book}
Liu, Z.; and Zhou, J. 2020.
\newblock Introduction to graph neural networks.
\newblock \emph{Synthesis Lectures on Artificial Intelligence and Machine
  Learning}, 14(2): 1--127.

\bibitem[{Miller(1995)}]{1995wordnet}
Miller, G.~A. 1995.
\newblock WordNet: A Lexical Database for English.
\newblock \emph{Commun. ACM}, 38(11): 39–41.

\bibitem[{Palmer(2009)}]{palmer2009semlink}
Palmer, M. 2009.
\newblock Semlink: Linking propbank, verbnet and framenet.
\newblock In \emph{Proceedings of the generative lexicon conference}, 9--15.
  GenLex-09, Pisa, Italy.

\bibitem[{Palmer, Gildea, and Kingsbury(2005)}]{palmer2005propbank}
Palmer, M.; Gildea, D.; and Kingsbury, P. 2005.
\newblock The proposition bank: An annotated corpus of semantic roles.
\newblock \emph{Computational linguistics}, 31(1): 71--106.

\bibitem[{Palmer, Kipper et~al.(2004)}]{palmer2004verbnet}
Palmer, M.; Kipper, K.~L.; et~al. 2004.
\newblock Verbnet.
\newblock \emph{The Oxford Handbook of Cognitive Science}.

\bibitem[{Riegel et~al.(2020)Riegel, Gray, Luus, Khan, Makondo, Akhalwaya,
  Qian, Fagin, Barahona, Sharma et~al.}]{riegel2020LNN}
Riegel, R.; Gray, A.; Luus, F.; Khan, N.; Makondo, N.; Akhalwaya, I.~Y.; Qian,
  H.; Fagin, R.; Barahona, F.; Sharma, U.; et~al. 2020.
\newblock Logical neural networks.
\newblock \emph{arXiv preprint arXiv:2006.13155}.

\bibitem[{Rogers, Kovaleva, and Rumshisky(2020)}]{rogers2020Bertology}
Rogers, A.; Kovaleva, O.; and Rumshisky, A. 2020.
\newblock A primer in bertology: What we know about how bert works.
\newblock \emph{Transactions of the Association for Computational Linguistics},
  8: 842--866.

\bibitem[{Sanchez-Lengeling et~al.(2021)Sanchez-Lengeling, Reif, Pearce, and
  Wiltschko}]{sanchez2021GentleGNN}
Sanchez-Lengeling, B.; Reif, E.; Pearce, A.; and Wiltschko, A.~B. 2021.
\newblock A gentle introduction to graph neural networks.
\newblock \emph{Distill}, 6(9): e33.

\bibitem[{Scarselli et~al.(2008)Scarselli, Gori, Tsoi, Hagenbuchner, and
  Monfardini}]{scarselli2008GNN}
Scarselli, F.; Gori, M.; Tsoi, A.~C.; Hagenbuchner, M.; and Monfardini, G.
  2008.
\newblock The graph neural network model.
\newblock \emph{IEEE transactions on neural networks}, 20(1): 61--80.

\bibitem[{Selvaraju et~al.(2017)Selvaraju, Cogswell, Das, Vedantam, Parikh, and
  Batra}]{selvaraju2017gradcam}
Selvaraju, R.~R.; Cogswell, M.; Das, A.; Vedantam, R.; Parikh, D.; and Batra,
  D. 2017.
\newblock Grad-cam: Visual explanations from deep networks via gradient-based
  localization.
\newblock In \emph{Proceedings of the IEEE international conference on computer
  vision}, 618--626.

\bibitem[{Serafini and Garcez(2016)}]{serafini2016LTN}
Serafini, L.; and Garcez, A.~d. 2016.
\newblock Logic tensor networks: Deep learning and logical reasoning from data
  and knowledge.
\newblock \emph{arXiv preprint arXiv:1606.04422}.

\bibitem[{Shi et~al.(2021)Shi, Lv, Seng, Zhang, Chen, and
  Xing}]{shi2021visualizingGCN}
Shi, X.; Lv, F.; Seng, D.; Zhang, J.; Chen, J.; and Xing, B. 2021.
\newblock Visualizing and understanding graph convolutional network.
\newblock \emph{Multimedia Tools and Applications}, 80(6): 8355--8375.

\bibitem[{Thomas et~al.(2019)Thomas, Heekeren, M{\"u}ller, and
  Samek}]{thomas2019neurimageDL}
Thomas, A.~W.; Heekeren, H.~R.; M{\"u}ller, K.-R.; and Samek, W. 2019.
\newblock Analyzing neuroimaging data through recurrent deep learning models.
\newblock \emph{Frontiers in neuroscience}, 13: 1321.

\bibitem[{Vaswani et~al.(2017)Vaswani, Shazeer, Parmar, Uszkoreit, Jones,
  Gomez, Kaiser, and Polosukhin}]{NIPS2017_attention}
Vaswani, A.; Shazeer, N.; Parmar, N.; Uszkoreit, J.; Jones, L.; Gomez, A.~N.;
  Kaiser, L.~u.; and Polosukhin, I. 2017.
\newblock Attention is All you Need.
\newblock In Guyon, I.; Luxburg, U.~V.; Bengio, S.; Wallach, H.; Fergus, R.;
  Vishwanathan, S.; and Garnett, R., eds., \emph{Advances in Neural Information
  Processing Systems}, volume~30. Curran Associates, Inc.

\bibitem[{Wu et~al.(2020)Wu, Pan, Chen, Long, Zhang, and
  Philip}]{wu2020gnnsurvey}
Wu, Z.; Pan, S.; Chen, F.; Long, G.; Zhang, C.; and Philip, S.~Y. 2020.
\newblock A comprehensive survey on graph neural networks.
\newblock \emph{IEEE transactions on neural networks and learning systems},
  32(1): 4--24.

\end{thebibliography}
\end{document}